\documentclass[final]{cvpr}

\usepackage{times}
\usepackage{epsfig}
\usepackage{graphicx}
\usepackage{amsmath}
\usepackage{amssymb}

\usepackage{booktabs}       
\usepackage{xcolor,colortbl}
\usepackage{bbm}

\definecolor{amber}{rgb}{1.0, 0.75, 0.0}

\usepackage[pagebackref=true,breaklinks=true,colorlinks,bookmarks=false]{hyperref}

\def\cvprPaperID{8522} 
\def\confYear{CVPR 2021}
\makeatletter
\def\@fnsymbol#1{\ensuremath{\ifcase#1\or \dagger\or \ddagger\or \mathsection\or \mathparagraph\or \|\or **\or \dagger\dagger \or \ddagger\ddagger \else\@ctrerr\fi}}
\makeatother

\begin{document}

\title{HOTR: End-to-End Human-Object Interaction Detection with Transformers}
\newcommand*\samethanks[1][\value{footnote}]{\footnotemark[#1]}
\author{
Bumsoo Kim\textsuperscript{\rm 1,2}\hspace{0.4cm}
Junhyun Lee\textsuperscript{\rm 2}\hspace{0.4cm}
Jaewoo Kang\textsuperscript{\rm 2}\hspace{0.4cm}
Eun-Sol Kim\textsuperscript{\rm 1}$^ ,$\thanks{corresponding authors}\hspace{0.4cm}
Hyunwoo J. Kim\textsuperscript{\rm 2}$^ ,$\samethanks\vspace{0.2cm}
\\\textsuperscript{\rm 1}Kakao Brain \hspace{0.4cm} \textsuperscript{\rm 2}Korea University
\\\tt\small \{bumsoo.brain, eunsol.kim\}@kakaobrain.com
\\\tt\small \{meliketoy, ljhyun33, kangj, hyunwoojkim\}@korea.ac.kr
}

\maketitle

\begin{abstract}
   Human-Object Interaction (HOI) detection is a task of identifying ``a set of interactions" in an image, which involves the i) localization of the subject (\ie, humans) and target (\ie, objects) of interaction, and ii) the classification of the interaction labels.
   Most existing methods have indirectly addressed this task by detecting human and object instances and individually inferring every pair of the detected instances.
   In this paper, we present a novel framework, referred by HOTR, which directly predicts a set of $\langle$human, object, interaction$\rangle$ triplets from an image based on a transformer encoder-decoder architecture.
   Through the set prediction, our method effectively exploits the inherent semantic relationships in an image and does not require time-consuming post-processing which is the main bottleneck of existing methods.
   Our proposed algorithm achieves the state-of-the-art performance in two HOI detection benchmarks with an inference time under 1 ms after object detection.
\end{abstract}

\section{Introduction}
\begin{figure}
    \centering
    \includegraphics[width=\columnwidth]{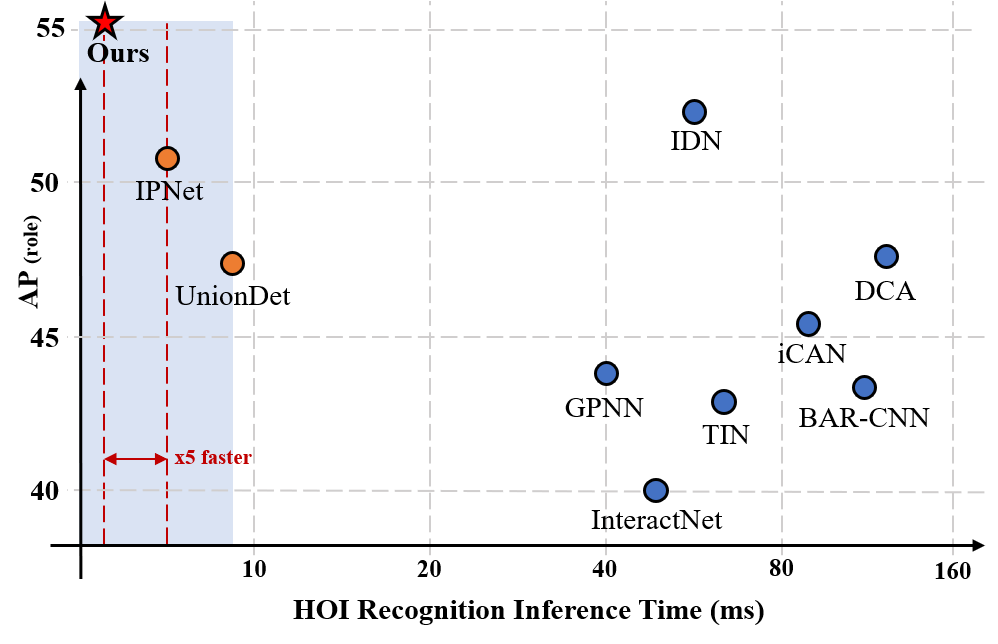}
    \caption{Time vs. Performance analysis for HOI detectors on V-COCO dataset. HOI recognition inference time is measured by subtracting the object detection time from the end-to-end inference time.
    {\color{blue}Blue circle} represents sequential HOI detectors, {\color{orange}orange circle} represents parallel HOI detectors and {\color{red}red star} represents ours.
    Our method achieves an HOI recognition inference time of 0.9ms, being significantly faster than the parallel HOI detectors such as IPNet~\cite{wang2020learning} or UnionDet~\cite{bkim2020uniondet} (the comparison between parallel HOI detectors is highlighted in blue).
    }
    \label{fig:fig_speed}
\end{figure}

Human-Object Interaction (HOI) detection has been formally defined in~\cite{gupta2015visual} as the task to predict a set of $\langle$human, object, interaction$\rangle$ triplets within an image.
Previous methods have addressed this task in an indirect manner by performing object detection first and associating $\langle$human, object$\rangle$ pairs afterward with separate post-processing steps.
Especially, early attempts (\ie, sequential HOI detectors~\cite{gao2018ican,li2019transferable,li2020pastanet,ulutan2020vsgnet}) have performed this association with a subsequent neural network, thus being time-consuming and computationally expensive.

To overcome the redundant inference structure of sequential HOI detectors, recent researches~\cite{wang2020learning,liao2020ppdm,bkim2020uniondet} proposed parallel HOI detectors.
These works explicitly localize interactions with either interaction boxes (\ie, the tightest box that covers both the center point of an object pair)~\cite{wang2020learning,liao2020ppdm} or union boxes (\ie, the tightest box that covers both the box regions of an object pair)~\cite{bkim2020uniondet}.
The localized interactions are associated with object detection results to complete the $\langle$human, object, interaction$\rangle$ triplet.
The time-consuming neural network inference is replaced with a simple matching based on heuristics such as distance~\cite{wang2020learning,liao2020ppdm} or IoU~\cite{bkim2020uniondet}.

However, previous works in HOI detection are still limited in two aspects;
i) They require additional post-processing steps like suppressing near-duplicate predictions and heuristic thresholding. 
ii) Although it has been shown that modeling relations between objects helps object detection~\cite{hu2018relation,carion2020end}, the effectiveness of considering high-level dependency for interactions in HOI detection has not yet been fully explored.

In this paper, we propose a fast and accurate HOI algorithm named HOTR (Human-Object interaction TRansformer) that predicts a set of human-object interactions in a scene at once with a direct set prediction approach.
We design an encoder-decoder architecture based on transformers to predict a set of HOI triplets, which enables the model to overcome both limitations of previous works.
First, direct set-level prediction enables us to eliminate hand-crafted post-processing stage.
Our model is trained in an end-to-end fashion with a set loss function that matches the predicted interactions with ground-truth $\langle$human, object, interaction$\rangle$ triplets.
Second, the self-attention mechanisms of transformers makes the model exploit the contextual relationships between human and object and their interactions, encouraging our set-level prediction framework more suitable for high-level scene understanding.

We evaluate our model in two HOI detection benchmarks: V-COCO and HICO-DET datasets.
Our proposed architecture achieves state-of-the-art performance on two datasets compared to both sequential and parallel HOI detectors.
Also, note that our method is much faster than other algorithms as illustrated in Figure \ref{fig:fig_speed}, by eliminating time-consuming post-processing through the direct set-level prediction.
The contribution of this work can be summarized as the following:
\begin{itemize}
    \item We propose HOTR, the first transformer-based set prediction approach in HOI detection. HOTR eliminates the hand-crafted post-processing stage of previous HOI detectors while being able to model the correlations between interactions.
    \item We propose various training and inference techniques for HOTR: HO Pointers to associate the outputs of two parallel decoders, a recomposition step to predict a set of final HOI triplets, and a new loss function to enable end-to-end training.
    \item HOTR achieves state-of-the-art performance on both benchmark datasets in HOI detection with an inference time under 1 ms, being significantly faster than previous parallel HOI detectors (5$\sim$9 ms).
\end{itemize}
\section{Related Work}
\subsection{Human-Object Interaction Detection}
\label{subsec:HOI}
Human-Object Interaction detection has been initially proposed in \cite{gupta2015visual}, and has been developed in two main streams: sequential methods and parallel methods.
In sequential methods, object detection is performed first and every pair of the detected object is inferred with a separate neural network to predict interactions.
Parallel HOI detectors perform object detection and interaction prediction in parallel and associates them with simple heuristics such as distance or IoU.
\newline

\noindent\textbf{\textit{Sequential HOI Detectors:}}
InteractNet~\cite{gkioxari2018detecting} extended an existing object detector by introducing an action-specific density map to localize target objects based on the human-centric appearance, and combined features from individual boxes to predict the interaction.
Note that interaction detection based on visual cues from individual boxes often suffers from the lack of contextual information.\newline
To this end, iCAN~\cite{gao2018ican} proposed an instance-centric attention module that extracts contextual features complementary to the features from the localized objects/humans.
No-Frills HOI detection~\cite{gupta2019no} propose a training and inference HOI detection pipeline only using simple multi-layer perceptron.
Graph-based approaches have proposed frameworks that can explicitly represent HOI structures with graphs~\cite{qi2018learning,ulutan2020vsgnet,gao2020drg,wang2020contextual,liu2020consnet}.
Deep Contextual Attention~\cite{wang2019deep} leverages contextual information by a contextual attention framework in HOI.
\cite{wang2020contextual} proposes a heterogeneous graph network that models humans and objects as different kinds of nodes.
Various external sources such as linguistic priors~\cite{peyre2019detecting,xu2019learning,li2020pastanet,gao2020drg,bansal2020detecting,zhong2020polysemy,liu2020amplifying} or human pose information~\cite{li2020detailed,zhou2019relation,li2019transferable,gupta2019no,wan2019pose,zhou2019relation} have also been leveraged for further improve performance.
Although sequential HOI detectors feature a fairly intuitive pipeline and solid performance, they are time-consuming and computationally expensive because of the additional neural network inference \textit{after} the object detection phase.\newline

\noindent\textbf{\textit{Parallel HOI Detectors:}}
Attempts for faster HOI detection has been also introduced in recent works as parallel HOI detectors.
These works have directly localized interactions with interaction points~\cite{wang2020learning,liao2020ppdm} or union boxes~\cite{bkim2020uniondet}, replacing the separate neural network for interaction prediction with a simple heuristic based matching with distance or IoUs.
Since they can be parallelized with existing object detectors, they feature fast inference time.
However, these works are limited in that they require a hand-crafted postprocessing stage to associate the localized interactions with object detection results.
This post-processing step i) requires manual search for the threshold, and ii) generates extra time complexity for matching each object pairs with the localized interactions (5$\sim$9 ms).

\subsection{Object Detection with Transformers}
\label{subsec:DETR}
DETR~\cite{carion2020end} has been recently proposed to eliminate the need for many hand-designed components in object detection while demonstrating good performance.
DETR infers a fixed-size set of $N$ predictions, in a single pass through the decoder, where $N$ is set to be significantly larger than the typical number of objects in an image.
The main loss for DETR produces an optimal bipartite matching between predicted and ground-truth objects.
Afterward, the object-specific losses (for class and bounding box) are optimized.
\begin{figure*}
    \centering
    \includegraphics[width=\textwidth]{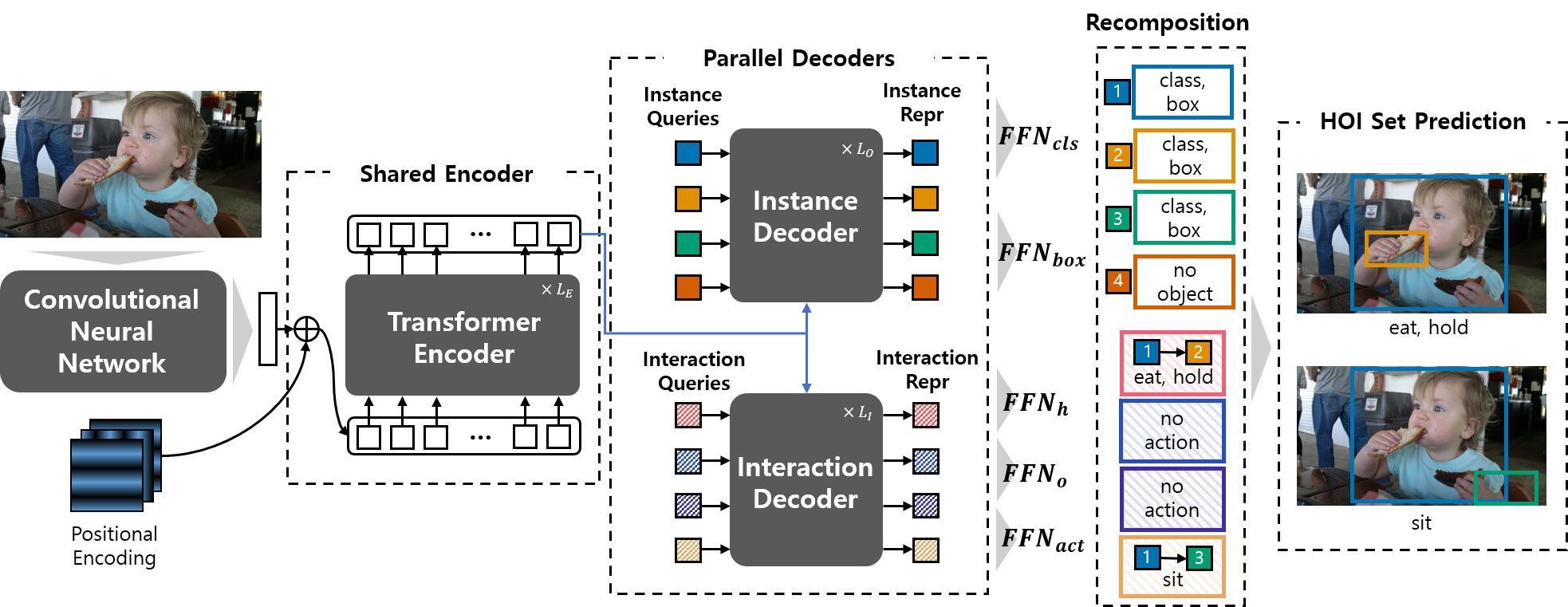}
    \caption{Overall pipeline of our proposed model. The Instance Decoder and Interaction Decoder run in parallel, and share the Encoder. In our recomposition, the interaction representations predicted by the Interaction Decoder are associated with the instance representations to predict a fixed set of HOI triplets (see Fig.\ref{fig:fig_nested}).
    The positional encoding is identical to~\cite{carion2020end}.} 
    \label{fig:fig_pipeline}
\end{figure*}
\section{Method}
The goal of this paper is to predict a set of $\langle$human, object, interaction$\rangle$ triplets while considering the inherent semantic relationships between the triplets in an end-to-end manner.
To achieve this goal, we formulate HOI detection as set prediction.
In this section, we first discuss the problems of directly extending the set prediction architecture for object detection~\cite{carion2020end} to HOI detection.
Then, we propose our architecture HOTR that parallelly predicts a set of object detection and associates the human and object of the interaction, while the self-attention in transformers models the relationships between the interactions.
Finally, we present the details of training for our model including Hungarian Matching for HOI detection and our loss function.

\subsection{Detection as Set Prediction}

We first start from object detection as set prediction with transformers, then show how we extend this architecture to capture HOI detection with transformers.
\newline

\noindent\textit{\textbf{Object Detection as Set Prediction.}}
Object Detection has been explored as a set prediction problem by DETR~\cite{carion2020end}.
Since object detection includes a single classification and a single localization for each object, the transformer encoder-decoder structure in DETR transforms $N$ positional embeddings to a set of $N$ predictions for the object class and bounding box.
\newline

\noindent\textit{\textbf{HOI Detection as Set Prediction.}}
Similar to object detection, HOI detection can be defined as a set prediction problem where each prediction includes the localization of a human region (\ie, \textit{subject} of the interaction), an object region (\ie, \textit{target} of the interaction) and multi-label classification of the interaction types.
One straightforward extension is to modify the MLP heads of DETR to transform each positional embedding to predict a human box, object box, and action classification.
However, this architecture poses a problem where the localization for the same object needs to be redundantly predicted with multiple positional embeddings (\eg, if the same person \textit{works on a computer} while \textit{sitting} on a chair, two different queries have to infer redundant regression for the same human).
\newline

\subsection{HOTR architecture}
The overall pipeline of HOTR is illustrated in Figure~\ref{fig:fig_pipeline}.
Our architecture features a transformer encoder-decoder structure with a shared encoder and two parallel decoders (\ie, instance decoder and interaction decoder).
The results of the two decoders are associated with using our proposed \textit{HO Pointers} to generate final HOI triplets.
We will introduce HO Pointers shortly after discussing the architecture of HOTR.
\newline

\begin{figure}
    \centering
    \includegraphics[width=\columnwidth]{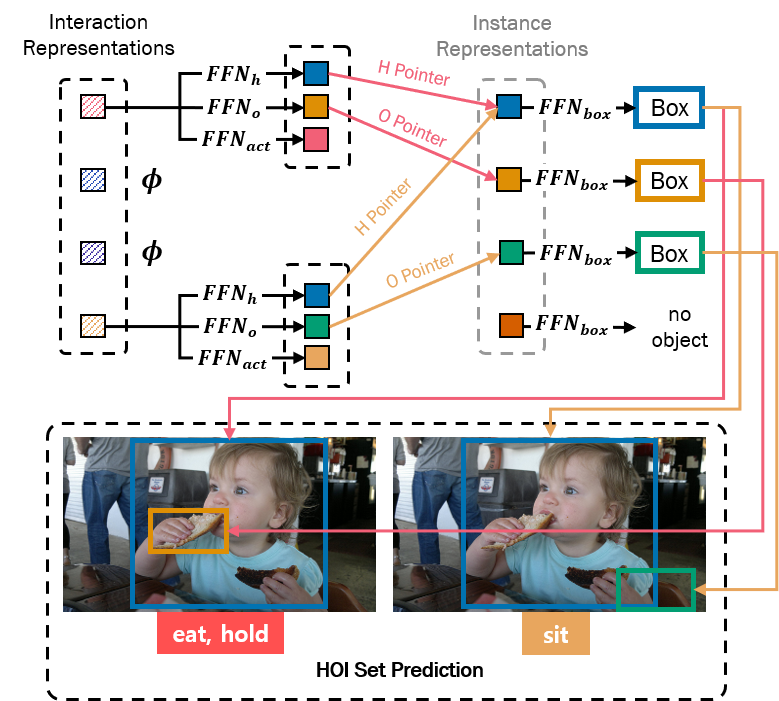}
    \caption{Conceptual illustration of how HO Pointers associates the interaction representations with instance representations. As instance representations are pre-trained to perform standard object detection, the interaction representation learns localization by predicting the \textit{pointer} to the index of the instance representations for each human and object boxes. Note that the index pointer prediction is obtained in parallel with instance representations.} 
    \label{fig:fig_nested}
\end{figure}

\noindent\textit{\textbf{Transformer Encoder-Decoder architecture.}}
Similar to DETR~\cite{carion2020end}, the global context is extracted from the input image by the backbone CNN and a shared encoder.
Afterward, two sets of positional embeddings (\ie, the instance queries and the interaction queries) are fed into the two parallel decoders (\ie, the instance decoder and interaction decoder in Fig.~\ref{fig:fig_pipeline}).
The instance decoder transforms the instance queries to instance representations for object detection while the interaction decoder transforms the interaction queries to interaction representations for interaction detection.
We apply feed-forward networks (FFNs) to the interaction representation and obtain a Human Pointer, an Object Pointer, and interaction type, see Fig. \ref{fig:fig_nested}.
In other words, the interaction representation localizes human and object regions by pointing the relevant instance representations using the Human Pointer and Object Pointer (HO Pointers), instead of directly regressing the bounding box.
Our architecture has several advantages compared to the direct regression approach.
We found that directly regressing the bounding box has a problem when an object participates in multiple interactions.
In the direct regression approach, the localization of the identical object differs across interactions.
Our architecture addresses this issue by having separate instance and interaction representations and associating them using HO Pointers.
Also, our architecture allows learning the localization more efficiently without the need of learning the localization redundantly for every interaction.
Note that our experiments show that our shared encoder is more effective to learn HO Pointers than two separate encoders.
\newline

\noindent\textit{\textbf{HO Pointers.}}
A conceptual overview of how HO Pointers associate the parallel predictions from the instance decoder and the interaction decoder is illustrated in Figure~\ref{fig:fig_nested}.
HO Pointers (\ie, Human Pointer and Object Pointer) contain the indices of the corresponding instance representations of the human and the object in the interaction.
After the interaction decoder transforms $K$ interaction queries to $K$ interaction representations, an  interaction representation  $z_i$ is fed into two feed-forward networks $\text{FFN}_h: \mathbb{R}^d\rightarrow \mathbb{R}^d$, $\text{FFN}_o: \mathbb{R}^d\rightarrow \mathbb{R}^d$ to obtain vectors $v_i^h$ and $v_i^o$, \ie, $v^h_i=\text{FFN}_h(z_i)$ and $v^o_i=\text{FFN}_o(z_i)$.
Then finally the Human/Object Pointers $\hat{c}^h_i$ and $\hat{c}^o_i$, which are the indices of the \textit{instance representations} with the highest similarity scores, are obtained by
\begin{equation}
\begin{split}
    \hat{c}^h_i=\mathop{\mathrm{argmax}}_{j}{\big(\text{sim}(v^h_i, \mu_j)\big)},\\
    \hat{c}^o_i=\mathop{\mathrm{argmax}}_{j}{\big(\text{sim}(v^o_i, \mu_j)\big)},  \\
\end{split}
\label{eq:eq_1}
\end{equation}
where $\mu_j$ is the $j$-th instance representation and sim$(u,v)=u^\top v/\|u\|\|v\|$.
\newline

\noindent\textit{\textbf{Recomposition for HOI Set Prediction.}}
From the previous steps, we now have the following:
i) $N$ instance representations $\mu$, and
ii) $K$ interaction representations $z$ and their HO Pointers $\hat{c}^h$ and $\hat{c}^o$.
Given $\gamma$ interaction classes, our recomposition is to apply the feed-forward networks for bounding box regression and action classification as $\text{FFN}_{\mbox{box}}: \mathbb{R}^d\rightarrow \mathbb{R}^4$, and $\text{FFN}_{\mbox{act}}: \mathbb{R}^d\rightarrow \mathbb{R}^{\gamma}$, respectively.
Then, the final HOI prediction for the $i$-th interaction representation $z_i$ is obtained by,
\begin{equation}
\begin{split}
    \hat{b}^h_i=&\quad\text{FFN}_{\mbox{box}}(\mu_{\hat{c}^h_i})\in\mathbb{R}^4,\\
    \hat{b}^o_i=&\quad\text{FFN}_{\mbox{box}}(\mu_{\hat{c}^o_i})\in\mathbb{R}^4,\\
    \hat{a}_i=&\quad\text{FFN}_{\mbox{act}}(z_i)\in\mathbb{R}^\gamma.
\end{split}
\label{eq:eq_2}
\end{equation}
The final HOI prediction by our HOTR is the set of $K$ triplets, $\{ \langle \hat{b}^h_i, \hat{b}_i^o, \hat{a}_i \rangle \}_{i=1}^K$.
\newline

\noindent\textit{\textbf{Complexity \& Inference time.}}
Previous parallel methods have substituted the costly pair-wise neural network inference with a fast matching of triplets (associating interaction regions with corresponding human regions and object regions based on distance~\cite{wang2020learning} or IoU~\cite{bkim2020uniondet}).
HOTR further reduces the inference time after object detection by associating $K$ interactions with $N$ instances, resulting in a smaller time complexity $\mathcal{O}(KN)$.
By eliminating the post-processing stages in the previous one-stage HOI detectors including NMS for the interaction region and triplet matching, HOTR diminishes the inference time by $4\sim 8$ms while showing improvement in performance.
\newline
\subsection{Training HOTR}
In this section, we explain the details of HOTR training.
We first introduce the cost matrix of Hungarian Matching for unique matching between the ground-truth HOI triplets and HOI set predictions obtained by recomposition.
Then, using the matching result, we define the loss for HO Pointers and the final training loss.
\newline

\noindent\textit{\textbf{Hungarian Matching for HOI Detection.}}
HOTR predicts $K$ HOI triplets that consist of human box, object box and binary classification for the $a$ types of actions.
Each prediction captures a unique $\langle$human,object$\rangle$ pair with one or more interactions.
$K$ is set to be larger than the typical number of interacting pairs in an image.
We start with the basic cost function that defines an optimal bipartite matching between predicted and ground truth HOI triplets, and then show how we modify this matching cost for our interaction representations.

Let $\mathcal{Y}$ denote the set of ground truth HOI triplets and $\hat{\mathcal{Y}}=\{\hat{y_i}\}_{i=1}^{K}$ as the set of $K$ predictions.
As $K$ is larger than the number of unique interacting pairs in the image, we consider $\mathcal{Y}$ also as a set of size $K$ padded with $\varnothing$ (no interaction).
To find a bipartite matching between these two sets we search for a permutation of $K$ elements $\sigma\in\mathfrak{S}_K$ with the lowest cost:
\begin{equation}
    \hat{\sigma}=\mathop{\mathrm{argmin}}_{\sigma\in\mathfrak{S}_K}\sum_{i}^{K}\mathcal{C}_{\text{match}}(y_i,\hat{y}_{\sigma(i)}),
\label{eq:eq_3}
\end{equation}
where $\mathcal{C}_{\text{match}}$ is a pair-wise \textit{matching cost} between ground truth $y_i$ and a prediction with index $\sigma(i)$.
However, since $y_i$ is in the form of $\langle$hbox,obox,action$\rangle$ and $\hat{y}_{\sigma(i)}$ is in the form of $\langle$hidx,oidx,action$\rangle$, we need to modify the cost function to compute the matching cost.

Let $\Phi:\text{idx}\rightarrow \text{box}$ be a mapping function from ground-truth $\langle$hidx,oidx$\rangle$ to ground-truth $\langle$hbox,obox$\rangle$ by optimal assignment for object detection.
Using the inverse mapping $\Phi^{-1}:\text{box}\rightarrow \text{idx}$, we get the ground-truth \textit{idx} from the ground-truth \textit{box}.

Let $M \in \mathbb{R}^{d \times N}$ be a set of normalized instance representations $\mu'=\mu/\|\mu\| \in \mathbb{R}^{d}$, \ie, $M=[\mu_1'  \ldots  \mu_N' ]$.
We compute $\hat{P}^h \in \mathbb{R}^{K\times N}$ that is the set of softmax predictions for the H Pointer in \eqref{eq:eq_1} given as
\begin{equation}
    \hat{P}^h=\|_{i=1}^K \mbox{softmax}((\bar{v}_i^h)^T M),
\end{equation}
where $\|_{i=1}^K$ denotes the vertical stack of row vectors and $\bar{v}_i^h=v_i^h/||v_i^h||$. $\hat{P}^o$ is analogously defined.

Given the ground-truth $y_{i} = (b_i^h,b_i^o,a_i), \hat{P}^h,\text{ and }, \hat{P}^o$, we convert the ground-truth box to indices by ${c}^h_i=\Phi^{-1}(b_i^h)$ and ${c}^o _i=\Phi^{-1}(b_i^o)$ and compute our matching cost function written as
\begin{equation}
\begin{split}
    \mathcal{C}_{\text{match}}(y_i, \hat{y}_{\sigma(i)})=-\alpha\cdot&\mathbbm{1}_{\{a_i\neq \varnothing\}}\hat{P}^{h}[\sigma(i),{c}^h_i] \\
    -\beta\cdot&\mathbbm{1}_{\{a_i\neq \varnothing\}}\hat{P}^{o}[\sigma(i),{c}^o_i] \\
    +&\mathbbm{1}_{\{a_i\neq\varnothing\}}\mathcal{L}_{\text{act}}(a_i,\hat{a}_{\sigma(i)}),
\end{split}
\end{equation}
where $\hat{P}[i,j]$ denotes the element at i-th row and j-th column, and $\hat{a}_{\sigma(i)}$ is the predicted action.
The action matching cost is calculated as  $\mathcal{L}_{\text{act}}(a_i,\hat{a}_{\sigma(i)})=\text{BCELoss}(a_i,\hat{a}_{\sigma(i)})$.
$\alpha$ and $\beta$ is set as a fixed number to balance the different scales of the cost function for index prediction and action classification.
\newline


\noindent\textit{\textbf{Final Set Prediction Loss for HOTR.}}
We then compute the Hungarian loss for all pairs matched above, where the loss for the HOI triplets has the localization loss and the action classification loss as
\begin{equation}
    \mathcal{L}_{\text{H}}=\sum_{i=1}^K\big[\mathcal{L}_{\text{loc}}(\text{c}^h_i, \text{c}^o_i, z_{\sigma(i)})+\mathcal{L}_{\text{act}}(a_i,\hat{a}_{\sigma(i)})\big].
\end{equation}
\newline
The localization loss $\mathcal{L}_{\text{loc}}(c^h_i, c^o_i, z_{\sigma(i)})$ is denoted as
\begin{equation}
\begin{split}
    \mathcal{L}_{\text{loc}}=&-\log\frac{\exp(\text{sim}(\text{FFN}_{h}(z_{\sigma(i)}),\mu_{c^h_i})/\tau)}{\sum_{k=1}^{N}\exp(\text{sim}(\text{FFN}_h(z_{\sigma(i)}),\mu_{k})/\tau)} \\
    & \\
    &-\log\frac{\exp(\text{sim}(\text{FFN}_{o}(z_{\sigma(i)}),\mu_{c^o_i}/\tau)}{\sum_{k=1}^{N}\exp(\text{sim}(\text{FFN}_{o}(z_{\sigma(i)}),\mu_k)/\tau)},
\end{split}
\end{equation}
where $\tau$ is the temperature that controls the smoothness of the loss function. We empirically found that $\tau=0.1$ is the best value for our experiments.
\newline

\noindent\textit{\textbf{Defining No-Interaction with HOTR.}}
In DETR~\cite{carion2020end}, maximizing the probability of the no-object class for the softmax output naturally suppresses the probability of other classes.
However, in HOI detection the action classification is a multi-label classification where each action is treated as an individual binary classification.
Due to the absence of an explicit class that can suppress the redundant predictions, HOTR ends up with multiple predictions for the same $\langle$human,object$\rangle$ pair.
Therefore, HOTR sets an explicit class that learns the \textit{interactiveness} (1 if there is \textit{any} interaction between the pair, 0 otherwise), and suppresses the predictions for redundant pairs that have a low interactiveness score (defined as No-Interaction class).
In our experiment in Table.~\ref{tab:Ablation}, we show that setting an explicit class for interactiveness contributes to the final performance.
\newline

\noindent\textit{\textbf{Implementation Details.}}
We train HOTR with AdamW~\cite{loshchilov2017decoupled}.
We set the transformer's initial learning rate to $10^{-4}$ and weight decay to $10^{-4}$.
All transformer weights are initialized with Xavier init~\cite{glorot2010understanding}.
For a fair evaluation with baselines, the Backbone, Encoder, and Instance Decoder are pre-trained in MS-COCO and frozen during training.
We use the scale augmentation as in DETR~\cite{carion2020end}, resizing the input images such that the shortest side is at least 480 and at most 800 pixels while the longest side at most is 1333.
\newline
\section{Experiments}
\label{sec:experiments}
In this section, we demonstrate the effectiveness of our model in HOI detection.
We first describe the two public datasets that we use as our benchmark: V-COCO and HICO-DET. 
Next, we show that HOTR successfully captures HOI triplets, by achieving state-of-the-art performance in both mAP and inference time.
Then, we provide a detailed ablation study of the HOTR architecture.

\subsection{Datasets}
To validate the performance of our model, we evaluate our model on two public benchmark datasets: the V-COCO (\textit{Verbs in COCO}) dataset and HICO-DET dataset.
\textbf{\textit{V-COCO}} is a subset of COCO and has 5,400 \texttt{trainval} images and 4,946 \texttt{test} images.
For V-COCO dataset, we report the $\text{AP}_{\text{role}}$ over $25$ interactions in two scenarios $\text{AP}_{\text{role}}^{\#1}$ and $\text{AP}_{\text{role}}^{\#2}$.
The two scenarios represent the different scoring ways for object occlusion cases. In Scenario1, the model should correctly predict the bounding box of the occluded object as [0,0,0,0] while predicting human bounding box and actions correctly.
In Scenario2, the model does not need to predict about the occluded object.
\textbf{\textit{HICO-DET}}~\cite{chao2018learning} is a subset of HICO dataset and has more than 150K annotated instances of human-object pairs in 47,051 images (37,536 training and 9,515 testing) and is annotated with 600 $\langle verb, object \rangle$ interaction types.
For HICO-DET, we report our performance in the \textit{Default} setting where we evaluate the detection on the full test set.
We follow the previous settings and report the mAP over three different category sets: (1) all 600 HOI categories in HICO (Full), (2) 138 HOI categories with less than 10 training instances (Rare), and (3) 462 HOI categories with 10 or more training instances (Non-Rare).
\subsection{Quantitative Analysis}
For quantitative analysis, we use the official evaluation code for computing the performance of both V-COCO and HICO-DET.
Table~\ref{tab:V-COCO role} and Table~\ref{tab:HICO-DET role} show the comparison of HOTR with the latest HOI detectors including both sequential and parallel methods.
For fair comparison, the instance detectors are fixed by the parameters pre-trained in MS-COCO.
All results in V-COCO dataset are evaluated with the fixed detector.
For the HICO-DET dataset, we provide both results using the fixed detector and the fine-tuned detector following the common evaluation protocol~\cite{bansal2020detecting,li2019transferable,hou2020visual,liu2020consnet,gao2020drg,li2020hoi,bkim2020uniondet,liao2020ppdm}.

Our HOTR achieves a new state-of-the-art performance on both V-COCO and HICO-DET datasets, while being the fastest parallel detector.
\begin{table}[h!]
  \centering
  \begin{tabular}{l|c|c c}
    \toprule
    Method & Backbone & $AP_{\text{role}}^{\#1}$ & $AP_{\text{role}}^{\#2}$ \\ \midrule\hline
    \multicolumn{4}{l}{\textit{Models with \textbf{external features}}} \\ \hline
    TIN (R$\text{P}_{\text{D}}\text{C}_{\text{D}}$)~\cite{li2019transferable} & R50 & 47.8 \\
    Verb Embedding~\cite{xu2019learning} & R50 & 45.9 \\
    RPNN~\cite{zhou2019relation} & R50 & - & 47.5 \\
    PMFNet~\cite{wan2019pose} & R50-FPN & 52.0 \\
    PastaNet~\cite{li2020pastanet} & R50-FPN & 51.0 & 57.5 \\
    PD-Net~\cite{zhong2020polysemy} & R50 & 52.0 & - \\
    ACP~\cite{kim2020detecting} & R152 & 53.0 & \\
    FCMNet~\cite{liu2020amplifying} & R50 & 53.1 & - \\
    ConsNet~\cite{liu2020consnet} & R50-FPN & 53.2 & - \\
    \hline
    \rowcolor[gray]{0.85}\multicolumn{4}{l}{\textit{\textbf{Sequential HOI Detectors}}} \\ \hline
    
    VSRL~\cite{gupta2015visual} & R50-FPN & 31.8 & - \\
    InteractNet~\cite{gkioxari2018detecting} & R50-FPN & 40.0 & 48.0 \\
    BAR-CNN~\cite{kolesnikov2019detecting} & R50-FPN & 43.6 & - \\
    GPNN~\cite{qi2018learning} & R152 & 44.0 & - \\
    iCAN~\cite{gao2018ican} & R50 & 45.3 & 52.4 \\
    TIN (R$\text{C}_{\text{D}}$)~\cite{li2019transferable} & R50 & 43.2 & - \\
    DCA~\cite{wang2019deep} & R50 & 47.3 & - \\
    VSGNet~\cite{ulutan2020vsgnet} & R152 & 51.8 & 57.0 \\
    VCL~\cite{hou2020visual} & R50-FPN & 48.3 & \\
    DRG~\cite{gao2020drg} & R50-FPN & 51.0 & \\
    IDN~\cite{li2020hoi} & R50 & 53.3 & 60.3 \\ \hline
    \rowcolor[gray]{0.85}\multicolumn{4}{l}{\textit{\textbf{Parallel HOI Detectors}}} \\ \hline
    
    IPNet~\cite{wang2020learning} & HG104 & 51.0 & - \\
    UnionDet~\cite{bkim2020uniondet} & R50-FPN & 47.5 & 56.2 \\ \midrule
    \textit{\textbf{Ours}} & R50 & \textbf{55.2} & \textbf{64.4} \\
    \bottomrule
  \end{tabular}
  \vspace{3pt}
  \caption{Comparison of performance on V-COCO test set. $AP_{\text{role}}^{\#1}$, $AP_{\text{role}}^{\#2}$ denotes the performance under Scenario1 and Scenario2 in V-COCO, respectively.}
  \label{tab:V-COCO role}
\end{table}
\begin{table*}[h!]
  \centering
  \small
  \begin{tabular}{l c c c c c c}
    \toprule
    \multicolumn{4}{c}{} & \multicolumn{3}{c}{\textbf{Default}} \\
    \cmidrule(r){5-7}
    Method & \hspace{12pt}Detector\hspace{12pt} & Backbone & \hspace{12pt}Feature\hspace{12pt} & \hspace{12pt}Full\hspace{12pt} & \hspace{6pt}Rare\hspace{6pt} & Non Rare\hspace{6pt} \\ \midrule\hline
    \rowcolor[gray]{0.85}\multicolumn{7}{l}{\textit{\textbf{Sequential HOI Detectors}}} \\ \hline
    \multicolumn{1}{l|}{InteractNet~\cite{gkioxari2018detecting}} & COCO & R50-FPN & \multicolumn{1}{c|}{A} & 9.94 & 7.16 & 10.77  \\
    \multicolumn{1}{l|}{GPNN~\cite{qi2018learning}} & COCO & R101 & \multicolumn{1}{c|}{A} & 13.11 & 9.41 & 14.23  \\
    \multicolumn{1}{l|}{iCAN~\cite{gao2018ican}} & COCO & R50 & \multicolumn{1}{c|}{A+S} & 14.84 & 10.45 & 16.15  \\
    \multicolumn{1}{l|}{DCA~\cite{wang2019deep}} & COCO & R50 & \multicolumn{1}{c|}{A+S} & 16.24 & 11.16 & 17.75  \\
    \multicolumn{1}{l|}{TIN~\cite{li2019transferable}} & COCO & R50 & \multicolumn{1}{c|}{A+S+P} & 17.03 & 13.42 & 18.11  \\
    \multicolumn{1}{l|}{RPNN~\cite{zhou2019relation}} & COCO & R50 & \multicolumn{1}{c|}{A+P} & 17.35 & 12.78 & 18.71  \\
    \multicolumn{1}{l|}{PMFNet~\cite{wan2019pose}} & COCO & R50-FPN & \multicolumn{1}{c|}{A+S+P} & 17.46 & 15.65 & 18.00  \\
    \multicolumn{1}{l|}{No-Frills HOI~\cite{gupta2019no}} & COCO & R152 & \multicolumn{1}{c|}{A+S+P} & 17.18 & 12.17 & 18.68  \\
    \multicolumn{1}{l|}{DRG~\cite{gao2020drg}} & COCO & R50-FPN & \multicolumn{1}{c|}{A+S+L} & 19.26 & 17.74 & 19.71  \\
    \multicolumn{1}{l|}{VCL~\cite{hou2020visual}} & COCO & R50 & \multicolumn{1}{c|}{A+S} & 19.43 & 16.55 & 20.29  \\
    \multicolumn{1}{l|}{VSGNet~\cite{ulutan2020vsgnet}} & COCO & R152 & \multicolumn{1}{c|}{A+S} & 19.80 & 16.05 & 20.91  \\
    \multicolumn{1}{l|}{FCMNet~\cite{liu2020amplifying}} & COCO & R50 & \multicolumn{1}{c|}{A+S+P} & 20.41	& 17.34	& 21.56  \\
    \multicolumn{1}{l|}{ACP~\cite{kim2020detecting}} & COCO & R152 & \multicolumn{1}{c|}{A+S+P} & 20.59 & 15.92 & 21.98  \\
    \multicolumn{1}{l|}{PD-Net~\cite{zhong2020polysemy}} & COCO & R50 & \multicolumn{1}{c|}{A+S+P+L} & 20.81 & 15.90 & 22.28  \\
    \multicolumn{1}{l|}{DJ-RN~\cite{li2020detailed}} & COCO & R50 & \multicolumn{1}{c|}{A+S+V} & 21.34 & 18.53 & 22.18  \\
    \multicolumn{1}{l|}{ConsNet~\cite{liu2020consnet}} & COCO & R50-FPN & \multicolumn{1}{c|}{A+S+L} & 22.15 & 17.12 & 23.65  \\
    \multicolumn{1}{l|}{PastaNet~\cite{li2020pastanet}} & COCO & R50 & \multicolumn{1}{c|}{A+S+P+L} & 22.65 & 21.17 & 23.09  \\
    \multicolumn{1}{l|}{IDN~\cite{li2020hoi}} & COCO & R50 & \multicolumn{1}{c|}{A+S} & 23.36 & \textbf{22.47} & 23.63  \\ \hline
    
    \multicolumn{1}{l|}{Functional Gen.~\cite{bansal2020detecting}} & HICO-DET & R101 & \multicolumn{1}{c|}{A+S+L} & 21.96 & 16.43 & 23.62  \\
    \multicolumn{1}{l|}{TIN~\cite{li2019transferable}} & HICO-DET & R50 & \multicolumn{1}{c|}{A+S+P} & 22.90 & 14.97 & 25.26  \\
    \multicolumn{1}{l|}{VCL~\cite{hou2020visual}} & HICO-DET & R50 & \multicolumn{1}{c|}{A+S} & 23.63 & 17.21 & 25.55  \\
    \multicolumn{1}{l|}{ConsNet~\cite{liu2020consnet}} & HICO-DET & R50-FPN & \multicolumn{1}{c|}{A+S+L} & 24.39 & 17.10 & 26.56  \\
    \multicolumn{1}{l|}{DRG~\cite{gao2020drg}} & HICO-DET & R50-FPN & \multicolumn{1}{c|}{A+S} & 24.53 & 19.47 & 26.04  \\
    \multicolumn{1}{l|}{IDN~\cite{li2020hoi}} & HICO-DET & R50 & \multicolumn{1}{c|}{A+S} & 24.58 & \textbf{20.33} & 25.86  \\ \midrule\hline
    
    \rowcolor[gray]{0.85}\multicolumn{7}{l}{\textit{\textbf{Parallel HOI Detectors}}} \\ \hline
    \multicolumn{1}{l|}{UnionDet~\cite{bkim2020uniondet}} & COCO & R50-FPN & \multicolumn{1}{c|}{A} & 14.25 & 10.23 & 15.46 \\
    \multicolumn{1}{l|}{IPNet~\cite{wang2020learning}} & COCO & R50-FPN & \multicolumn{1}{c|}{A} & 19.56 & 12.79 & 21.58 \\
    \multicolumn{1}{l|}{\textbf{\textit{Ours}}} & COCO & R50 & \multicolumn{1}{c|}{A} & \textbf{23.46} & 16.21 & \textbf{25.62} \\ \hline
    
    \multicolumn{1}{l|}{UnionDet~\cite{bkim2020uniondet}} & HICO-DET & R50-FPN & \multicolumn{1}{c|}{A} & 17.58 & 11.72 & 19.33 \\
    \multicolumn{1}{l|}{PPDM~\cite{liao2020ppdm}} & HICO-DET & HG104 & \multicolumn{1}{c|}{A} & 21.10 & 14.46 & 23.09 \\
    \multicolumn{1}{l|}{\textbf{\textit{Ours}}} & HICO-DET & R50 & \multicolumn{1}{c|}{A} & \textbf{25.10} & 17.34 & \textbf{27.42} \\
    \bottomrule
  \end{tabular}
  \vspace{3pt}
  \caption{Performance comparison in HICO-DET. The Detector column is denoted as `COCO' for the models that freeze the object detectors with the weights pre-trained in MS-COCO and `HICO-DET' if the object detector is fine-tuned with the HICO-DET train set. The each letter in Feature column stands for A: Appearance (Visual features), S: Interaction Patterns (Spatial Correlations~\cite{gao2018ican}), P: Pose Estimation, L: Linguistic Priors, V: Volume~\cite{li2020detailed}.}
  \label{tab:HICO-DET role}
\end{table*}
Table~\ref{tab:V-COCO role} shows our result in the V-COCO dataset with both Scenario1 and Scenario2.
HOTR outperforms the state-of-the-art parallel HOI detector~\cite{wang2020learning} in Scenario1 with a margin of $4.2$mAP.
Table~\ref{tab:HICO-DET role} shows the result in HICO-DET in the Default setting for each Full/Rare/Non-Rare class.
Due to the noisy labeling for objects in the HICO-DET dataset, fine-tuning the pre-trained object detector on the HICO-DET train set provides a prior that benefits the overall performance~\cite{bansal2020detecting}.
Therefore, we evaluate our performance in HICO-DET dataset under two conditions: i) using pre-trained weights from MS-COCO which are frozen during training (denoted as COCO in the Detector column) and ii) performance after fine-tuning the pre-trained detector on the HICO-DET train set (denoted as HICO-DET in the Detector column).
Our model outperforms the state-of-the-art parallel HOI detector under both conditions by a margin of $4.1$mAP and $4$mAP, respectively.
Below, we provide a more detailed analysis of our performance.
\newline

\noindent\textit{\textbf{HOTR vs Sequential Prediction.}}
In comparative analysis with various HOI methods summarized in Table~\ref{tab:V-COCO role} and \ref{tab:HICO-DET role}, we also compare the experimental results of HOTR with sequential prediction methods.
Even though the sequential methods take advantages from additional information while HOTR only utilize visual information, HOTR outperforms the state-of-the-art sequential HOI detector~\cite{li2020hoi} in both Scenario1 and Scenario2 by $1.9$ mAP and $4.1$ mAP in V-COCO while showing comparable performance (with a margin of 0.1$\sim$0.52 mAP) in the Default(Full) evaluation of HICO-DET.
\newline

\noindent\textit{\textbf{Performance on HICO-DET Rare Categories.}}
HOTR shows state-of-the-art performance across both sequential and parallel HOI detectors in the Full evaluation for HICO-DET dataset (see Table.~\ref{tab:HICO-DET role}).
However, HOTR underperforms than baseline methods~\cite{li2020hoi} in the Rare setting.
Since this setting deals with the action categories that has less than 10 training instances, it is difficult to achieve accuracy on this setting without the help of external features.
Therefore, most of the studies that have shown high performance in Rare settings make use of additional information, such as spatial layouts~\cite{gao2018ican}, pose information~\cite{li2019transferable}, linguistic priors~\cite{li2020pastanet}, and coherence patterns between the humans and objects~\cite{li2020hoi}.
In this work, our method is a completely vision-based pipeline but if we include the prior knowledge, we expect further improvement in the Rare setting.

\noindent\textit{\textbf{Time analysis.}}
Since the inference time of the object detector network (e.g., Faster-RCNN~\cite{ren2015faster}) can vary  depending on benchmark settings (e.g., the library, CUDA, CUDNN version or hyperparameters), the time analysis is based on the pure inference time of the HOI interaction prediction model excluding the time of the object detection phase for fair comparison with our model.
For detailed analysis, HOTR takes an average of $36.3$ms for the backbone and encoder, $23.8$ms for the instance decoder and interaction decoder (note that the two decoders run in parallel), and $0.9$ms for the recomposition and final HOI triplet inference.
We excluded the i/o times in all models including the time of previous models loading the RoI align features of Faster-RCNN (see Figure.\ref{fig:fig_speed} for a speed vs time comparison).
Note that our HOTR runs $\times5\sim\times9$ faster compared to the state-of-the-art parallel HOI detectors, since an explicit post-processing stage to assemble the detected objects and interaction regions is replaced with a simple $O(KN)$ search to infer the HO Pointers.
\newline
\subsection{Ablation Study}
\begin{table}[h!]
  \centering
  \begin{tabular}{l c c}
    \toprule
    Method & $AP_{\text{role}}^{\#1}$ & Default(Full) \\ \midrule
    HOTR & \textbf{55.2} & \textbf{23.5} \\
    w/o HO Pointers & 39.3 & 17.2 \\
    w/o Shared Encoders & 33.9 & 14.5 \\
    w/o Interactiveness Suppression & 52.2 & 22.0 \\
\bottomrule
\end{tabular}
\vspace{3pt}
\caption{Ablation Study on both V-COCO test set (scenario 1, $AP_{\text{role}}^{\#1}$) and HICO-DET test set (Default, Full setting without fine-tuning the object detector)}
\label{tab:Ablation}
\end{table}
In this section, we explore how each of the components of HOTR contributes to the final performance.
Table~\ref{tab:Ablation} shows the final performance in the V-COCO test set after excluding each components of HOTR.
We perform all experiments with the most basic R50-C4 backbone, and fix the transformer layers to 6 and attention heads 8 and the feed-forward network dimension to $d=1024$ unless otherwise mentioned.
\newline

\noindent\textit{\textbf{With vs Without HO Pointers.}}
In HOTR, the interaction representation localizes human and object region by pointing the relevant instance representations using the Human Pointer and Object Pointer (HO Pointers), instead of directly regressing the bounding box.
We pose that our architecture has advantages compared to the direct regression approach, since directly regressing the bounding box for every interaction prediction requires redundant bounding box regression for the same object when an object participates in multiple interactions.
Based on the performance gap ($55.2\rightarrow 39.3$ in V-COCO and $23.5\rightarrow 17.2$ in HICO-DET), it can be concluded that using HO Pointers alleviates the issue of direct regression approach.
\newline

\noindent\textit{\textbf{Shared Encoder vs Separate Encoders.}}
From the Fig. 2, the architecture having separate encoders for each Instance and Interaction Decoder can be considered.
In this ablation, we verify the role of the shared encoder of the HOTR.
In Table~\ref{tab:Ablation}, it is shown that sharing the encoder outperforms the model with separate encoders by a margin of $21.3$mAP and $9.0$mAP in V-COCO and HICO-DET, respectively.
We suppose the reason is that the shared encoder helps the decoders learn common visual patterns, thus the HO Pointers can share the overall context.
\newline

\noindent\textit{\textbf{With vs Without Interactiveness Suppression.}}
Unlike softmax based classification where maximizing the probability for the no-object class can explicitly diminish the probability of other classes, action classification is a multi-label binary classification that treats each class independently.
So HOTR sets an explicit class that learns the \textit{interactiveness}, and suppresses the predictions for redundant pairs that have low probability.
Table~\ref{tab:Ablation} shows that setting an explicit class for interactiveness contributes $3$mAP to the final performance.
\newline

\section{Conclusion}
\label{sec:conclusion}
In this paper, we present HOTR, the first transformer-based set prediction approach in human-object interaction problem.
The set prediction approach of HOTR eliminates the hand-crafted post-processing steps of previous HOI detectors while being able to model the correlations between interactions.
We propose various training and inference techniques for HOTR: HOI decomposition with parallel decoders for training, recomposition layer based on similarity for inference, and interactiveness suppression.
We develop a novel set-based matching for HOI detection that associates the interaction representations to point at instance representations.
Our model achieves state-of-the-art performance in two benchmark datasets in HOI detection: V-COCO and HICO-DET, with a significant margin to previous parallel HOI detectors.
HOTR achieves state-of-the-art performance on both benchmark datasets in HOI detection with an inference time under 1 ms, being significantly faster than previous parallel HOI detectors (5$\sim$9 ms).
\\

\noindent\textbf{Acknowledgments.}
{\small This research was partly supported by the Institute of Information \& communications Technology Planning \& Evaluation (IITP) grants funded by the Korea government (MSIT) (No.2021-0-00025, Development of Integrated Cognitive Drone AI for Disaster/Emergency Situations), (IITP-2021-0-01819, the ICT Creative Consilience program), and National Research Foundation of Korea (NRF2020R1A2C3010638, NRF-2016M3A9A7916996).}
\newpage

{\small
\bibliographystyle{ieee_fullname}
\bibliography{cvpr}

\begin{thebibliography}{10}\itemsep=-1pt

\bibitem{bansal2020detecting}
Ankan Bansal, Sai~Saketh Rambhatla, Abhinav Shrivastava, and Rama Chellappa.
\newblock Detecting human-object interactions via functional generalization.
\newblock In {\em AAAI}, pages 10460--10469, 2020.

\bibitem{carion2020end}
Nicolas Carion, Francisco Massa, Gabriel Synnaeve, Nicolas Usunier, Alexander
  Kirillov, and Sergey Zagoruyko.
\newblock End-to-end object detection with transformers.
\newblock {\em arXiv preprint arXiv:2005.12872}, 2020.

\bibitem{chao2018learning}
Yu-Wei Chao, Yunfan Liu, Xieyang Liu, Huayi Zeng, and Jia Deng.
\newblock Learning to detect human-object interactions.
\newblock In {\em 2018 ieee winter conference on applications of computer
  vision (wacv)}, pages 381--389. IEEE, 2018.

\bibitem{gao2020drg}
Chen Gao, Jiarui Xu, Yuliang Zou, and Jia-Bin Huang.
\newblock Drg: Dual relation graph for human-object interaction detection.
\newblock In {\em European Conference on Computer Vision}, pages 696--712.
  Springer, 2020.

\bibitem{gao2018ican}
Chen Gao, Yuliang Zou, and Jia-Bin Huang.
\newblock ican: Instance-centric attention network for human-object interaction
  detection.
\newblock {\em arXiv preprint arXiv:1808.10437}, 2018.

\bibitem{gkioxari2018detecting}
Georgia Gkioxari, Ross Girshick, Piotr Doll{\'a}r, and Kaiming He.
\newblock Detecting and recognizing human-object interactions.
\newblock In {\em Proceedings of the IEEE Conference on Computer Vision and
  Pattern Recognition}, pages 8359--8367, 2018.

\bibitem{glorot2010understanding}
Xavier Glorot and Yoshua Bengio.
\newblock Understanding the difficulty of training deep feedforward neural
  networks.
\newblock In {\em Proceedings of the thirteenth international conference on
  artificial intelligence and statistics}, pages 249--256, 2010.

\bibitem{gupta2015visual}
Jitendra Gupta, Saurabh~Malik.
\newblock Visual semantic role labeling.
\newblock {\em arXiv preprint arXiv:1505.04474}, 2015.

\bibitem{gupta2019no}
Tanmay Gupta, Alexander Schwing, and Derek Hoiem.
\newblock No-frills human-object interaction detection: Factorization, layout
  encodings, and training techniques.
\newblock In {\em Proceedings of the IEEE International Conference on Computer
  Vision}, pages 9677--9685, 2019.

\bibitem{hou2020visual}
Zhi Hou, Xiaojiang Peng, Yu Qiao, and Dacheng Tao.
\newblock Visual compositional learning for human-object interaction detection.
\newblock {\em arXiv preprint arXiv:2007.12407}, 2020.

\bibitem{hu2018relation}
Han Hu, Jiayuan Gu, Zheng Zhang, Jifeng Dai, and Yichen Wei.
\newblock Relation networks for object detection.
\newblock In {\em Proceedings of the IEEE Conference on Computer Vision and
  Pattern Recognition}, pages 3588--3597, 2018.

\bibitem{bkim2020uniondet}
Bumsoo Kim, Taeho Choi, Jaewoo Kang, and Hyunwoo Kim.
\newblock Uniondet: Union-level detection towards real-time human-object
  interaction detection.
\newblock In {\em Proceedings of the European conference on computer vision
  (ECCV)}, 2020.

\bibitem{kim2020detecting}
Dong-Jin Kim, Xiao Sun, Jinsoo Choi, Stephen Lin, and In~So Kweon.
\newblock Detecting human-object interactions with action co-occurrence priors.
\newblock {\em arXiv preprint arXiv:2007.08728}, 2020.

\bibitem{kolesnikov2019detecting}
Alexander Kolesnikov, Alina Kuznetsova, Christoph Lampert, and Vittorio
  Ferrari.
\newblock Detecting visual relationships using box attention.
\newblock In {\em Proceedings of the IEEE International Conference on Computer
  Vision Workshops}, pages 0--0, 2019.

\bibitem{li2020detailed}
Yong-Lu Li, Xinpeng Liu, Han Lu, Shiyi Wang, Junqi Liu, Jiefeng Li, and Cewu
  Lu.
\newblock Detailed 2d-3d joint representation for human-object interaction.
\newblock In {\em Proceedings of the IEEE/CVF Conference on Computer Vision and
  Pattern Recognition}, pages 10166--10175, 2020.

\bibitem{li2020hoi}
Yong-Lu Li, Xinpeng Liu, Xiaoqian Wu, Yizhuo Li, and Cewu Lu.
\newblock Hoi analysis: Integrating and decomposing human-object interaction.
\newblock {\em Advances in Neural Information Processing Systems}, 33, 2020.

\bibitem{li2020pastanet}
Yong-Lu Li, Liang Xu, Xinpeng Liu, Xijie Huang, Yue Xu, Shiyi Wang, Hao-Shu
  Fang, Ze Ma, Mingyang Chen, and Cewu Lu.
\newblock Pastanet: Toward human activity knowledge engine.
\newblock In {\em Proceedings of the IEEE/CVF Conference on Computer Vision and
  Pattern Recognition}, pages 382--391, 2020.

\bibitem{li2019transferable}
Yong-Lu Li, Siyuan Zhou, Xijie Huang, Liang Xu, Ze Ma, Hao-Shu Fang, Yanfeng
  Wang, and Cewu Lu.
\newblock Transferable interactiveness knowledge for human-object interaction
  detection.
\newblock In {\em Proceedings of the IEEE Conference on Computer Vision and
  Pattern Recognition}, pages 3585--3594, 2019.

\bibitem{liao2020ppdm}
Yue Liao, Si Liu, Fei Wang, Yanjie Chen, Chen Qian, and Jiashi Feng.
\newblock Ppdm: Parallel point detection and matching for real-time
  human-object interaction detection.
\newblock In {\em Proceedings of the IEEE/CVF Conference on Computer Vision and
  Pattern Recognition}, pages 482--490, 2020.

\bibitem{liu2020amplifying}
Y Liu, Q Chen, and A Zisserman.
\newblock Amplifying key cues for human-object-interaction detection.
\newblock {\em Lecture Notes in Computer Science}, 2020.

\bibitem{liu2020consnet}
Ye Liu, Junsong Yuan, and Chang~Wen Chen.
\newblock Consnet: Learning consistency graph for zero-shot human-object
  interaction detection.
\newblock In {\em Proceedings of the 28th ACM International Conference on
  Multimedia}, pages 4235--4243, 2020.

\bibitem{loshchilov2017decoupled}
Ilya Loshchilov and Frank Hutter.
\newblock Decoupled weight decay regularization.
\newblock {\em arXiv preprint arXiv:1711.05101}, 2017.

\bibitem{peyre2019detecting}
Julia Peyre, Ivan Laptev, Cordelia Schmid, and Josef Sivic.
\newblock Detecting unseen visual relations using analogies.
\newblock In {\em Proceedings of the IEEE International Conference on Computer
  Vision}, pages 1981--1990, 2019.

\bibitem{qi2018learning}
Siyuan Qi, Wenguan Wang, Baoxiong Jia, Jianbing Shen, and Song-Chun Zhu.
\newblock Learning human-object interactions by graph parsing neural networks.
\newblock In {\em Proceedings of the European Conference on Computer Vision
  (ECCV)}, pages 401--417, 2018.

\bibitem{ren2015faster}
Shaoqing Ren, Kaiming He, Ross Girshick, and Jian Sun.
\newblock Faster r-cnn: Towards real-time object detection with region proposal
  networks.
\newblock In {\em Advances in neural information processing systems}, pages
  91--99, 2015.

\bibitem{ulutan2020vsgnet}
Oytun Ulutan, ASM Iftekhar, and Bangalore~S Manjunath.
\newblock Vsgnet: Spatial attention network for detecting human object
  interactions using graph convolutions.
\newblock In {\em Proceedings of the IEEE/CVF Conference on Computer Vision and
  Pattern Recognition}, pages 13617--13626, 2020.

\bibitem{wan2019pose}
Bo Wan, Desen Zhou, Yongfei Liu, Rongjie Li, and Xuming He.
\newblock Pose-aware multi-level feature network for human object interaction
  detection.
\newblock In {\em Proceedings of the IEEE International Conference on Computer
  Vision}, pages 9469--9478, 2019.

\bibitem{wang2020contextual}
Hai Wang, Wei-shi Zheng, and Ling Yingbiao.
\newblock Contextual heterogeneous graph network for human-object interaction
  detection.
\newblock {\em arXiv preprint arXiv:2010.10001}, 2020.

\bibitem{wang2019deep}
Tiancai Wang, Rao~Muhammad Anwer, Muhammad~Haris Khan, Fahad~Shahbaz Khan,
  Yanwei Pang, Ling Shao, and Jorma Laaksonen.
\newblock Deep contextual attention for human-object interaction detection.
\newblock {\em arXiv preprint arXiv:1910.07721}, 2019.

\bibitem{wang2020learning}
Tiancai Wang, Tong Yang, Martin Danelljan, Fahad~Shahbaz Khan, Xiangyu Zhang,
  and Jian Sun.
\newblock Learning human-object interaction detection using interaction points.
\newblock In {\em Proceedings of the IEEE/CVF Conference on Computer Vision and
  Pattern Recognition}, pages 4116--4125, 2020.

\bibitem{xu2019learning}
Bingjie Xu, Yongkang Wong, Junnan Li, Qi Zhao, and Mohan~S Kankanhalli.
\newblock Learning to detect human-object interactions with knowledge.
\newblock In {\em Proceedings of the IEEE Conference on Computer Vision and
  Pattern Recognition}, 2019.

\bibitem{zhong2020polysemy}
Xubin Zhong, Changxing Ding, Xian Qu, and Dacheng Tao.
\newblock Polysemy deciphering network for human-object interaction detection.
\newblock In {\em Proc. Eur. Conf. Comput. Vis}, 2020.

\bibitem{zhou2019relation}
Penghao Zhou and Mingmin Chi.
\newblock Relation parsing neural network for human-object interaction
  detection.
\newblock In {\em Proceedings of the IEEE International Conference on Computer
  Vision}, pages 843--851, 2019.

\end{thebibliography}
}

\end{document}